\documentclass{article}

\usepackage[dblblindworkshop,final]{neurips_2026}
\workshoptitle{Foundation Models for Temporal Systems: From Forecasting to World Modeling (FMTS 2026)}
\makeatletter\renewcommand{\@noticestring}{}\makeatother

\usepackage[utf8]{inputenc}
\usepackage[T1]{fontenc}
\usepackage{hyperref}
\usepackage{url}
\usepackage{booktabs}
\usepackage{amsmath}
\usepackage{amsfonts}
\usepackage{graphicx}
\usepackage{placeins}
\usepackage{float}
\usepackage{nicefrac}
\usepackage{microtype}
\usepackage[table]{xcolor}
\definecolor{cellblue}{RGB}{100,135,225}
\usepackage{enumitem}
\usepackage{etoc}
\usepackage{array}
\usepackage{longtable}
\newcolumntype{L}[1]{>{\raggedright\arraybackslash}p{#1}}
\providecommand{\Description}[1]{}

\title{From Forecasting to World Modeling:\\Can Temporal Foundation Models Recover Causal Response?}

\author{%
  Juwon Hong \\
  Seoul National University \\
  Seoul, South Korea \\
  \texttt{juwonhong@snu.ac.kr} \\
  \And
  Minha Hwang \\
  Microsoft \\
  San Francisco, USA \\
  \texttt{minha.hwang@gmail.com} \\
  \And
  Venkatesh Shankar \\
  Cox School of Business \\
  Southern Methodist University \\
  Dallas, Texas, USA \\
  \texttt{vshankar@smu.edu}
}

\begin{document}

\maketitle
\etocdepthtag.toc{mtmain}

\begin{abstract}
A model that forecasts sales accurately can still mislead a retailer about how sales respond to a price
change. We evaluate released tabular and
time-series foundation models, as well as a foundation model pretrained specifically to estimate causal
effects. We report three findings. First, forecast
accuracy does not indicate the accuracy of the predicted effect of price, both on simulated data in which
the true effect is known and on real retail sales records. Second, for the tabular foundation models,
additional context improves forecasts but not the predicted effect of price.
Third, instrumental variables, the standard remedy in econometrics, reduce this error only when supplied in
a specific form and increase it for every model under confounding when supplied as raw instruments.
\end{abstract}

\section{Introduction}
\label{sec:intro}

A model that forecasts sales accurately can still mislead a retailer about the consequences of changing a
price. Retailers rely on sales forecasts to set prices, yet the two tasks pose different questions. A
forecast predicts the sales that typically accompanied the observed inputs in past data, whereas a pricing
decision requires the sales that would follow a change in price with all other inputs held fixed. The two
answers diverge because firms set prices with expected demand in mind \citep{villasboas1999,
nevo2001}; a retailer might, for instance, discount more deeply when it anticipates strong demand. Past prices therefore
reflect demand conditions that the data do not record. A forecasting model reduces its error by exploiting
this association but thereby credits price with part of the effect of demand.

We show that released temporal foundation models are vulnerable to this divergence. We evaluate the tabular
foundation models TabPFN \citep{tabpfn2}, TabICLv2 \citep{tabicl} and TabFM \citep{tabfm}, the
time-series foundation model Chronos-2 \citep{chronos2}, and the causal foundation model CCPFN \citep{ccpfn}. Causal foundation models might be expected to avoid the problem, since they are pretrained specifically to
predict the outcomes of interventions \citep{causalpfn, dopfn}. The pretraining data of such models as CausalPFN \citep{causalpfn, cfmsurvey} and CCPFN \citep{ccpfn}, however, are generated under the assumption that the model inputs include every factor that influences both price and sales. Retail data violate this assumption because prices
depend on demand that the data do not record. Their pretraining therefore does not prepare them to separate the effect of price from the association between price and sales. We find that additional data improve forecasts
without improving the predicted effect of price, and that instrumental variables, the standard remedy in
econometrics, help only when supplied in a specific form.

\section{Prediction, Causal Recovery and Evaluation}
\label{sec:setup}

\paragraph{Prediction and causal recovery.} A forecasting model predicts an outcome from the inputs recorded
with the outcome. Here the outcome is $q$, the units of a product that a store sells in a week. We denote the
price by $p$ and the remaining inputs, such as the product, the store, the season and promotions, by $x$. A
model that minimizes prediction error learns the conditional mean $\mathbb{E}[q \mid p, x]$, the sales that
typically accompanied the given inputs in the data. Causal
recovery instead concerns an input that an agent controls, here the price that the retailer sets. The
question is what the outcome would be if the agent set the input to a new value and left all other
conditions unchanged. We write the answer as $\mathbb{E}[q \mid \mathrm{do}(p), x]$. The two quantities
differ when the agent sets the input in response to an unobserved factor that also affects the outcome.

\paragraph{Confounding and instruments.} Such a factor is called a confounder. An observed confounder can be included as a model input, which makes
the conditional mean equal to the causal quantity; an unobserved confounder cannot, and the two remain
different. In retail, the retailer sets the price from the observed inputs $x$ and from unobserved
conditions $\xi$, such as anticipated demand. The same $\xi$ also affects sales:
\begin{equation}
p = f(x, \delta\xi), \qquad q = g(p, x, \xi) + \varepsilon.
\label{eq:dgp}
\end{equation}
The parameter $\delta$ sets how strongly price responds to $\xi$. Because $\xi$ enters both equations, the
observed price carries information about $\xi$. A linear example illustrates the consequence. Let
${\log q = \beta \log p + \xi + \varepsilon}$ and ${\log p = -\delta\xi + \nu}$ with $\delta \ge 0$,
where $\beta$ is the true effect of price. A model fitted to the resulting data learns the slope
\begin{equation}
\beta - \delta\,\frac{\operatorname{Var}(\xi)}{\operatorname{Var}(\log p)}.
\label{eq:bias}
\end{equation}
The fitted slope forecasts well, because low prices did coincide with high sales in the data. The same
slope, however, overstates how much sales would fall after a price increase, and additional data from the same
pricing behavior do not close the gap. The standard remedy in econometrics is an instrumental variable, or instrument: a variable $z$ that
shifts price, is independent of $\xi$, and affects sales only through price \citep{berry1994, blp1995}.
In Equation~\eqref{eq:dgp}, $z$ would enter $f$ but not $g$. An instrument thus provides price variation that
is independent of $\xi$. The response of sales to this variation reflects the effect of price alone and
therefore identifies $\beta$.

\paragraph{The test environment.} Evaluating whether a model recovers $\beta$ requires a control market that
is identical except for the absence of confounding, ground-truth sales under each price intervention,
$\mathbb{E}[q \mid \mathrm{do}(p), x]$, and valid
instruments. The simulated data of CausalDemand~\citep{causaldemand}, calibrated to
real store records, satisfy all three (Appendix~\ref{app:data}). First, each dataset comes in two versions
that share every random draw and differ only in $\delta$ (Appendix~\ref{sec:steps}): prices respond to $\xi$ in the confounded version
($\delta = 0.30$) and not in the unconfounded one ($\delta = 0$). Second, the simulator records $\xi$ for
every observation. The simulator obtains ground-truth sales under an intervention by replacing $p$ with a
new price $p'$ and re-evaluating $g(p', x, \xi)$ with all other terms unchanged. Third, the benchmark provides
valid instruments by construction.

\paragraph{The test.} Each model is trained once and then evaluated on the same held-out product-store-weeks
in two tasks (Appendix~\ref{app:mechanism}). Forecasting evaluates $\mathbb{E}[q \mid p, x]$: the model predicts
sales $\hat q$ at the observed prices, and the prediction is compared with the expected sales $q$.
Counterfactual prediction evaluates $\mathbb{E}[q \mid \mathrm{do}(p), x]$: the price of one product or brand
is raised or lowered by 10\% with all other inputs unchanged (Appendix~\ref{sec:scenario-prices}), the model predicts sales $\hat q^{\mathrm{cf}}$
at the new prices, and its predicted ratio of sales after and before the change, $\hat q^{\mathrm{cf}}/\hat q$,
is compared with the true ratio from the simulator, $q^{\mathrm{cf}}/q$. A model is thus scored on how sales
respond to the price change, not on their level, which the forecast error already measures. The benchmark defines 32 such interventions, which differ in the target product or brand and in
the direction of the change, so that the evaluation covers the market as a whole rather than a single price.
Both tasks are scored by the weighted mean absolute percentage error (WMAPE), with the counterfactual error
averaged over the 32 interventions (Appendix~\ref{app:metric} gives the exact definitions):
\begin{equation}
\text{Forecast error} = \frac{\sum |\hat q - q|}{\sum q}, \qquad
\text{Counterfactual error} = \frac{\sum q\,\big|\hat q^{\mathrm{cf}}/\hat q - q^{\mathrm{cf}}/q\big|}{\sum q^{\mathrm{cf}}}.
\label{eq:wmape}
\end{equation}

\section{Results}
\label{sec:results}

We compare the foundation models (TabPFN, TabICLv2, TabFM, CCPFN and Chronos-2) with LightGBM and with double machine learning (Double ML)
\citep{chernozhukov2018}, an econometric method that can use instruments.

\begin{table}[t]
  \centering
  \footnotesize
  \setlength{\tabcolsep}{3pt}
  \caption{\textbf{On real retail data, the most accurate forecasters often predict the wrong direction of the
  price effect.} Own-sign accuracy is the share of rows on which the model predicts that sales fall after a
  10\% price increase; a correct model scores 1. Appendix~\ref{sec:fmts-table1} documents the complete setup.}
  \label{tab:actual}
  \begin{tabular}{lcccc}
    \toprule
    \textbf{Category} & \multicolumn{2}{c}{Snack crackers} & \multicolumn{2}{c}{Breakfast cereals} \\
    \cmidrule(lr){2-3} \cmidrule(lr){4-5}
    \textbf{Metric} & Forecast WMAPE & Own-sign accuracy & Forecast WMAPE & Own-sign accuracy \\
    \midrule
    Chronos-2 & 0.368\,[0.357,\,0.381] & 0.834\,[0.806,\,0.865] & 0.499\,[0.472,\,0.531] & 0.942\,[0.931,\,0.952] \\
    TabPFN & 0.383\,[0.370,\,0.396] & 0.844\,[0.825,\,0.863] & 0.380\,[0.371,\,0.389] & 0.788\,[0.774,\,0.801] \\
    TabICLv2 & 0.437\,[0.425,\,0.450] & 0.852\,[0.823,\,0.878] & 0.413\,[0.403,\,0.423] & 0.892\,[0.877,\,0.907] \\
    TabFM & 0.484\,[0.468,\,0.501] & 0.844\,[0.824,\,0.864] & 0.494\,[0.477,\,0.512] & 0.869\,[0.859,\,0.878] \\
    LightGBM & 0.389\,[0.380,\,0.398] & 0.915\,[0.894,\,0.935] & 0.367\,[0.359,\,0.375] & 0.838\,[0.820,\,0.853] \\
    Double ML & 0.463\,[0.447,\,0.478] & 1.000\,[1.000,\,1.000] & 0.442\,[0.432,\,0.452] & 1.000\,[1.000,\,1.000] \\
    \bottomrule
  \end{tabular}
\end{table}

\paragraph{Experiment 1.} Before turning to the simulated data, we examine whether the
gap between forecasting and causal recovery appears in real sales records. Real records do not contain sales under
a price intervention, $\mathbb{E}[q \mid \mathrm{do}(p), x]$, so the counterfactual error cannot be computed. We therefore check only the direction
of the predicted response: after a 10\% price increase, a correct model predicts lower sales. The models with the lowest forecast error predict unchanged or higher sales on a substantial
share of rows, whereas Double ML, which uses
instruments, has higher forecast error but predicts lower
sales on every row (Table~\ref{tab:actual}). Low forecast error thus does not ensure that a model responds
to price in the correct direction.

\begin{figure}[!b]
  \centering
  \includegraphics[width=0.8\linewidth]{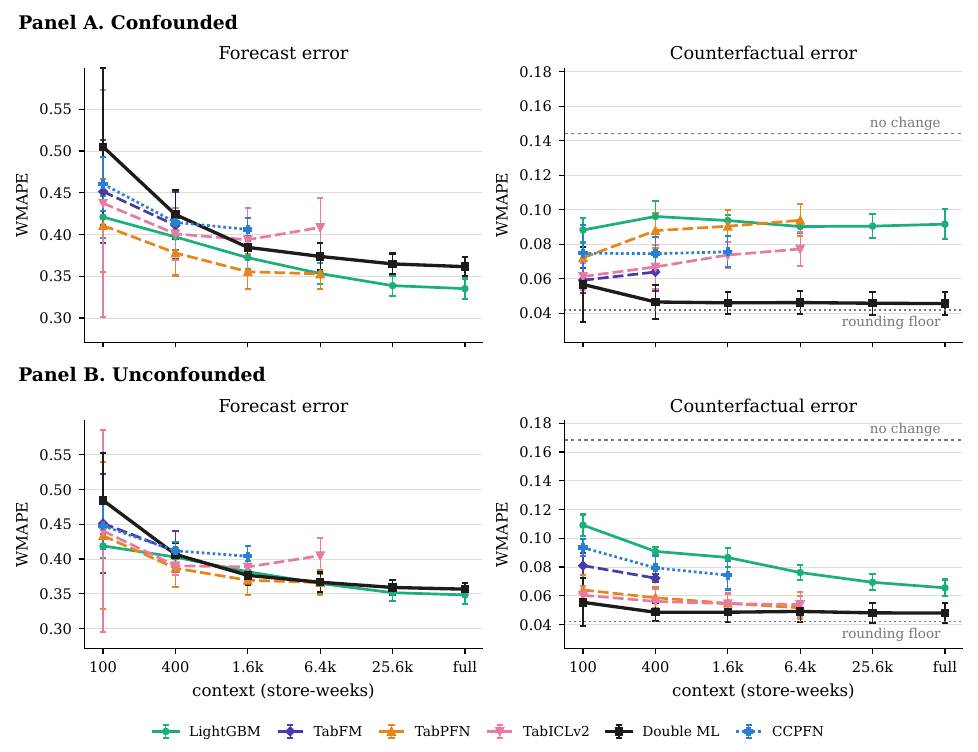}
  \caption{\textbf{More data improve forecasts, but counterfactuals only without confounding.} Panel~A uses the confounded data ($\delta = 0.30$) and Panel~B the unconfounded data
  ($\delta = 0$). The horizontal axis is the amount of context in store-weeks; \emph{full} is the whole
  panel (102,340 store-weeks). TabFM accepts at most 400 store-weeks, and TabPFN and TabICLv2 at most 6,400. CCPFN is run up to 1,600 store-weeks.
  The gray lines mark the error of predicting no change (0.144 in Panel~A, 0.168 in Panel~B) and the rounding
  floor (0.042) that remains for a model that knows the true demand. Points are means over five datasets with 95\% confidence intervals. Appendix~\ref{sec:fmts-fig1} documents the complete setup of this experiment.}
  \label{fig:divergence}
\end{figure}

\paragraph{Experiment 2.} We vary the amount of context that each
model receives, as training data for LightGBM and Double ML and as in-context examples for the foundation
models. The foundation models are evaluated
with their released weights and without fine-tuning, and each receives the price change in the input format
it supports (Appendix~\ref{sec:fmts-inputs}). Under
confounding, additional data generally reduce forecast error but leave the counterfactual error of every model without instruments flat or
increasing. The contrast matches
Equation~\eqref{eq:bias}: more data from the same pricing behavior sharpen the estimate of
$\mathbb{E}[q \mid p, x]$ but not of $\mathbb{E}[q \mid \mathrm{do}(p), x]$. Without confounding, additional data generally reduce both errors.

\begin{table}[!b]
  \centering
  \footnotesize
  \setlength{\tabcolsep}{3pt}
  \caption{\textbf{Instruments reduce counterfactual error as residuals but increase it as raw instruments.} Entries are means over five datasets with 95\% confidence intervals. Double ML uses the instruments directly in its estimation and therefore has a single entry. Darker cells mark lower error within each column of a panel. Appendix~\ref{sec:fmts-table2} documents the complete setup of this
  experiment.}
  \label{tab:results}
  \begin{tabular}{lccccccc}
    \toprule
    & \multicolumn{3}{c}{Forecast WMAPE} & & \multicolumn{3}{c}{Counterfactual WMAPE} \\
    \cmidrule(lr){2-4} \cmidrule(lr){6-8}
    & no instruments & residuals & raw instrument & & no instruments & residuals & raw instrument \\
    \midrule
    \multicolumn{8}{l}{\textbf{Panel A. Confounded}} \\
    LightGBM & \cellcolor{cellblue!60}0.335\,$\pm$\,0.012 & \cellcolor{cellblue!60}0.301\,$\pm$\,0.018 & \cellcolor{cellblue!60}0.306\,$\pm$\,0.020 &  & \cellcolor{cellblue!4}0.092\,$\pm$\,0.009 & \cellcolor{cellblue!34}0.064\,$\pm$\,0.005 & \cellcolor{cellblue!23}0.153\,$\pm$\,0.024 \\
    TabICLv2 & \cellcolor{cellblue!5}0.409\,$\pm$\,0.035 & \cellcolor{cellblue!5}0.381\,$\pm$\,0.013 & \cellcolor{cellblue!31}0.354\,$\pm$\,0.014 &  & \cellcolor{cellblue!34}0.077\,$\pm$\,0.010 & \cellcolor{cellblue!60}0.052\,$\pm$\,0.009 & \cellcolor{cellblue!8}0.179\,$\pm$\,0.023 \\
    TabPFN & \cellcolor{cellblue!47}0.353\,$\pm$\,0.018 & \cellcolor{cellblue!44}0.325\,$\pm$\,0.016 & \cellcolor{cellblue!46}0.329\,$\pm$\,0.019 &  & 0.094\,$\pm$\,0.009 & \cellcolor{cellblue!60}0.052\,$\pm$\,0.009 & 0.193\,$\pm$\,0.014 \\
    TabFM & \cellcolor{cellblue!3}0.411\,$\pm$\,0.040 & \cellcolor{cellblue!16}0.365\,$\pm$\,0.031 & \cellcolor{cellblue!26}0.363\,$\pm$\,0.043 &  & \cellcolor{cellblue!60}0.064\,$\pm$\,0.011 & \cellcolor{cellblue!45}0.059\,$\pm$\,0.007 & \cellcolor{cellblue!40}0.123\,$\pm$\,0.043 \\
    CCPFN & 0.415\,$\pm$\,0.011 & 0.389\,$\pm$\,0.014 & 0.407\,$\pm$\,0.018 &  & \cellcolor{cellblue!38}0.075\,$\pm$\,0.009 & 0.080\,$\pm$\,0.012 & \cellcolor{cellblue!37}0.128\,$\pm$\,0.024 \\
    Chronos-2 & \cellcolor{cellblue!43}0.358\,$\pm$\,0.013 & \cellcolor{cellblue!31}0.344\,$\pm$\,0.015 & \cellcolor{cellblue!8}0.393\,$\pm$\,0.013 &  & \cellcolor{cellblue!22}0.083\,$\pm$\,0.002 & \cellcolor{cellblue!2}0.079\,$\pm$\,0.003 & \cellcolor{cellblue!60}0.088\,$\pm$\,0.004 \\
    Double ML & \multicolumn{3}{c}{0.362\,$\pm$\,0.012} & & \multicolumn{3}{c}{0.046\,$\pm$\,0.007} \\
    \midrule
    \multicolumn{8}{l}{\textbf{Panel B. Unconfounded}} \\
    LightGBM & \cellcolor{cellblue!60}0.348\,$\pm$\,0.013 & \cellcolor{cellblue!60}0.348\,$\pm$\,0.012 & \cellcolor{cellblue!60}0.348\,$\pm$\,0.013 &  & \cellcolor{cellblue!41}0.066\,$\pm$\,0.006 & \cellcolor{cellblue!44}0.066\,$\pm$\,0.006 & \cellcolor{cellblue!44}0.071\,$\pm$\,0.007 \\
    TabICLv2 & \cellcolor{cellblue!7}0.405\,$\pm$\,0.025 & 0.410\,$\pm$\,0.016 & \cellcolor{cellblue!26}0.391\,$\pm$\,0.012 &  & \cellcolor{cellblue!57}0.054\,$\pm$\,0.009 & \cellcolor{cellblue!57}0.054\,$\pm$\,0.008 & \cellcolor{cellblue!60}0.053\,$\pm$\,0.007 \\
    TabPFN & \cellcolor{cellblue!43}0.366\,$\pm$\,0.019 & \cellcolor{cellblue!43}0.366\,$\pm$\,0.017 & \cellcolor{cellblue!41}0.372\,$\pm$\,0.022 &  & \cellcolor{cellblue!60}0.052\,$\pm$\,0.008 & \cellcolor{cellblue!60}0.051\,$\pm$\,0.008 & \cellcolor{cellblue!49}0.065\,$\pm$\,0.005 \\
    TabFM & \cellcolor{cellblue!1}0.411\,$\pm$\,0.029 & \cellcolor{cellblue!11}0.399\,$\pm$\,0.037 & \cellcolor{cellblue!24}0.393\,$\pm$\,0.030 &  & \cellcolor{cellblue!33}0.072\,$\pm$\,0.003 & \cellcolor{cellblue!45}0.065\,$\pm$\,0.004 & \cellcolor{cellblue!39}0.076\,$\pm$\,0.006 \\
    CCPFN & 0.412\,$\pm$\,0.012 & \cellcolor{cellblue!3}0.407\,$\pm$\,0.013 & 0.424\,$\pm$\,0.017 &  & \cellcolor{cellblue!24}0.079\,$\pm$\,0.009 & \cellcolor{cellblue!23}0.086\,$\pm$\,0.012 & \cellcolor{cellblue!27}0.090\,$\pm$\,0.017 \\
    Chronos-2 & \cellcolor{cellblue!50}0.359\,$\pm$\,0.014 & \cellcolor{cellblue!40}0.369\,$\pm$\,0.015 & \cellcolor{cellblue!38}0.376\,$\pm$\,0.010 &  & 0.097\,$\pm$\,0.005 & 0.108\,$\pm$\,0.006 & 0.120\,$\pm$\,0.008 \\
    Double ML & \multicolumn{3}{c}{0.357\,$\pm$\,0.009} & & \multicolumn{3}{c}{0.048\,$\pm$\,0.007} \\

    \bottomrule
  \end{tabular}
\end{table}

\paragraph{Experiment 3.} We supply the instruments to
each model in three forms that differ only in the input columns added: none, control-function residuals,
and the raw instruments (Table~\ref{tab:results}). The residuals come from a regression of price on $z$ and the other inputs and carry the part of price correlated with $\xi$ \citep{petrintrain2010, wooldridge2015}. Under
confounding, the residuals reduce counterfactual error for most models, whereas the raw instruments
increase counterfactual error for every model. For LightGBM, TabICLv2 and TabPFN, the error with raw instruments (0.153 to 0.193) exceeds that of predicting no change (0.144). For the tabular models, however, the raw instruments forecast better than no
instruments. Choosing by forecast error would therefore select the raw instruments, which predict the effect of
a price change less accurately.

\section{Conclusion}
\label{sec:conclusion}

Our results show that accurate forecasts do not imply an accurate counterfactual effect of price. Additional data
improved the causal recovery only when prices did not respond to demand, and a foundation model pretrained
for causal inference predicted the effect of price no more accurately than the forecasting models. Instrumental variables mostly reduced
the error when supplied as the part of price that they do not explain, whereas raw instruments
increased the error under confounding while improving forecast accuracy for the tabular models. One limitation is that the models differ in training objective, capacity and pretraining; our comparison
characterizes currently available models under
their standard use rather than the capabilities of their architectures. Within this scope, evaluations of temporal foundation models as world models
should measure the accuracy of predicted responses to interventions alongside forecast accuracy,
because forecast accuracy alone can favor the models that answer such questions least accurately.

\FloatBarrier

\section*{Acknowledgments}

Dominick's data courtesy of the James M. Kilts Center for Marketing, University of Chicago Booth School of
Business.

\bibliographystyle{plainnat}
\bibliography{references}

@inproceedings{dopfn,
  title     = {{Do-PFN}: In-Context Learning for Causal Effect Estimation},
  author    = {Robertson, Jake and Reuter, Arik and Guo, Siyuan and Hollmann, Noah and Hutter, Frank and Sch\"{o}lkopf, Bernhard},
  booktitle = {Advances in Neural Information Processing Systems},
  editor    = {Belgrave, D. and Zhang, C. and Lin, H. and Pascanu, R. and Koniusz, P. and Ghassemi, M. and Chen, N.},
  volume    = {38},
  pages     = {174811--174848},
  publisher = {Curran Associates, Inc.},
  year      = {2025},
  doi       = {10.52202/085713-5814}
}

@inproceedings{causalpfn,
  title     = {{CausalPFN}: Amortized Causal Effect Estimation via In-Context Learning},
  author    = {Balazadeh Meresht, Vahid and Kamkari, Hamidreza and Thomas, Valentin and Ma, Junwei and Li, Bingru and Cresswell, Jesse and Krishnan, Rahul},
  booktitle = {Advances in Neural Information Processing Systems},
  editor    = {Belgrave, D. and Zhang, C. and Lin, H. and Pascanu, R. and Koniusz, P. and Ghassemi, M. and Chen, N.},
  volume    = {38},
  pages     = {154945--154984},
  publisher = {Curran Associates, Inc.},
  year      = {2025},
  doi       = {10.52202/085713-5184}
}

@misc{cfmsurvey,
      title={Causal Foundation Models}, 
      author={Christopher Stith and Hossein Rahmani and Jesse C. Cresswell},
      year={2026},
      eprint={2609.03003},
      archivePrefix={arXiv},
      primaryClass={cs.LG},
      url={https://arxiv.org/abs/2609.03003}, 
}

@misc{chronos2,
      title={Chronos-2: From Univariate to Universal Forecasting}, 
      author={Abdul Fatir Ansari and Oleksandr Shchur and Jaris K\"{u}ken and Andreas Auer and Boran Han and Pedro Mercado and Syama Sundar Rangapuram and Huibin Shen and Lorenzo Stella and Xiyuan Zhang and Mononito Goswami and Shubham Kapoor and Danielle C. Maddix and Pablo Guerron and Tony Hu and Junming Yin and Nick Erickson and Prateek Mutalik Desai and Hao Wang and Huzefa Rangwala and George Karypis and Yuyang Wang and Michael Bohlke-Schneider},
      year={2025},
      eprint={2510.15821},
      archivePrefix={arXiv},
      primaryClass={cs.LG},
      url={https://arxiv.org/abs/2510.15821}, 
}

@inproceedings{tabicl,
  title     = {{TabICLv2}: A Better, Faster, Scalable, and Open Tabular Foundation Model},
  author    = {Qu, Jingang and Holzm\"{u}ller, David and Varoquaux, Ga\"{e}l and Le Morvan, Marine},
  booktitle = {Proceedings of the 43rd International Conference on Machine Learning},
  series    = {Proceedings of Machine Learning Research},
  volume    = {306},
  pages     = {102477--102555},
  publisher = {PMLR},
  year      = {2026},
  url       = {https://proceedings.mlr.press/v306/qu26a.html}
}

@article{tabpfn2,
  title     = {Accurate predictions on small data with a tabular foundation model},
  author    = {Hollmann, Noah and M{\"u}ller, Samuel and Purucker, Lennart and
               Krishnakumar, Arjun and K{\"o}rfer, Max and Hoo, Shi Bin and
               Schirrmeister, Robin Tibor and Hutter, Frank},
  journal   = {Nature},
  volume    = {637},
  number    = {8045},
  pages     = {319--326},
  year      = {2025},
  publisher = {Nature Publishing Group},
  doi       = {10.1038/s41586-024-08328-6}
}

@article{berry1994,
  title   = {Estimating Discrete-Choice Models of Product Differentiation},
  author  = {Berry, Steven T.},
  journal = {The RAND Journal of Economics},
  volume  = {25},
  number  = {2},
  pages   = {242--262},
  year    = {1994},
  doi={10.2307/2555829}
}

@article{blp1995,
  title   = {Automobile Prices in Market Equilibrium},
  author  = {Berry, Steven and Levinsohn, James and Pakes, Ariel},
  journal = {Econometrica},
  volume  = {63},
  number  = {4},
  pages   = {841--890},
  year    = {1995},
  doi={10.2307/2171802}
}

@article{nevo2001,
  title   = {Measuring Market Power in the Ready-to-Eat Cereal Industry},
  author  = {Nevo, Aviv},
  journal = {Econometrica},
  volume  = {69},
  number  = {2},
  pages   = {307--342},
  year    = {2001},
  doi={10.1111/1468-0262.00194}
}

@article{petrintrain2010,
  title   = {A Control Function Approach to Endogeneity in Consumer Choice Models},
  author  = {Petrin, Amil and Train, Kenneth},
  journal = {Journal of Marketing Research},
  volume  = {47},
  number  = {1},
  pages   = {3--13},
  year    = {2010},
  doi={10.1509/jmkr.47.1.3}
}

@article{wooldridge2015,
  title   = {Control Function Methods in Applied Econometrics},
  author  = {Wooldridge, Jeffrey M.},
  journal = {Journal of Human Resources},
  volume  = {50},
  number  = {2},
  pages   = {420--445},
  year    = {2015},
  doi={10.3368/jhr.50.2.420}
}

@article{chernozhukov2018,
  title   = {Double/Debiased Machine Learning for Treatment and Structural Parameters},
  author  = {Chernozhukov, Victor and Chetverikov, Denis and Demirer, Mert and
             Duflo, Esther and Hansen, Christian and Newey, Whitney and Robins, James},
  journal = {The Econometrics Journal},
  volume  = {21},
  number  = {1},
  pages   = {C1--C68},
  year    = {2018},
  doi={10.1111/ectj.12097}
}

@article{villasboas1999,
  title   = {Endogeneity in Brand Choice Models},
  author  = {Villas-Boas, J. Miguel and Winer, Russell S.},
  journal = {Management Science},
  volume  = {45},
  number  = {10},
  pages   = {1324--1338},
  year    = {1999},
  doi={10.1287/mnsc.45.10.1324}
}

@unpublished{causaldemand,
  title  = {{CausalDemand}: Benchmarking Causal Recovery of Demand from Retail
            Transactions and Product Text},
  author = {Hong, Juwon and Hwang, Minha and Shankar, Venkatesh},
  year   = {2026},
  note   = {Working paper}
}

@misc{ccpfn,
  title         = {Causal Foundation Models with Continuous Treatments},
  author        = {Stith, Christopher and Barath, Medha and Balazadeh, Vahid and Cresswell, Jesse C. and Krishnan, Rahul G.},
  year          = {2026},
  eprint        = {2605.15133},
  archivePrefix = {arXiv},
  primaryClass  = {cs.LG},
  url           = {https://arxiv.org/abs/2605.15133},
  note          = {ICML 2026 Workshop on Foundation Models for Structured Data}
}

@misc{tabfm,
  title        = {Introducing {TabFM}: A Zero-Shot Foundation Model for Tabular Data},
  author       = {Kong, Weihao and Das, Abhimanyu},
  year         = {2026},
  month        = jun,
  howpublished = {Google Research Blog},
  url          = {https://research.google/blog/introducing-tabfm-a-zero-shot-foundation-model-for-tabular-data/}
}

@article{magnolfi2025,
  author  = {Magnolfi, Lorenzo and McClure, Jonathon and Sorensen, Alan},
  title   = {Triplet Embeddings for Demand Estimation},
  journal = {American Economic Journal: Microeconomics},
  volume  = {17}, number = {1}, pages = {282--307}, year = {2025},
  doi     = {10.1257/mic.20220248}}

@article{tellis1988,
  author  = {Tellis, Gerard J.},
  title   = {The Price Elasticity of Selective Demand: A Meta-Analysis of
             Econometric Models of Sales},
  journal = {Journal of Marketing Research},
  volume  = {25}, number = {4}, pages = {331--341}, year = {1988},
  doi     = {10.1177/002224378802500401}}

\appendix
\clearpage
\raggedbottom
\etocdepthtag.toc{mtappendix}
\etocsettagdepth{mtmain}{none}
\etocsettagdepth{mtappendix}{subsection}
\renewcommand{\contentsname}{Appendix}
\tableofcontents
\clearpage

\section{The Simulated Data}\label{sec:dgp}\label{app:data}

\begin{figure}[H]
\centering
\includegraphics[width=\textwidth]{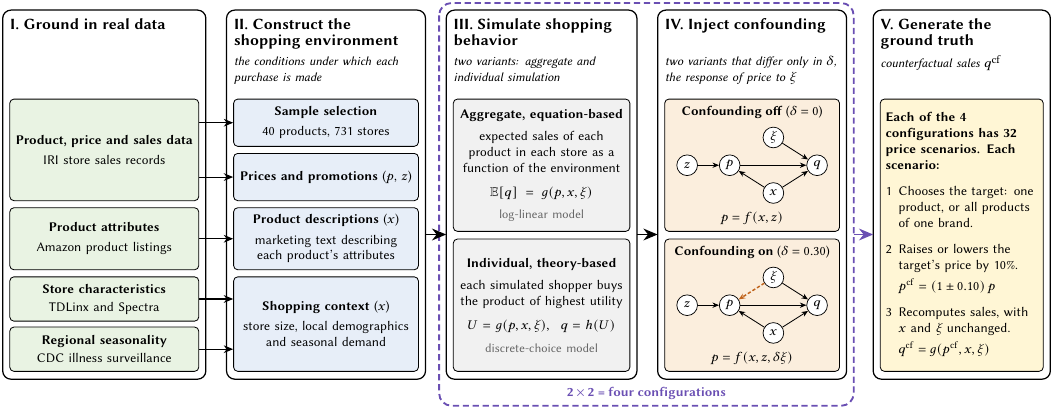}
\caption{Overview of how the CausalDemand datasets are generated.}
\Description{A five-panel pipeline diagram. Panel I lists four real data sources: IRI store sales
records, Amazon product listings, TDLinx and Spectra store profiles, and CDC illness surveillance.
Arrows lead to panel II, the shopping environment: sample selection of 40 products and 731 stores,
prices and promotions, product descriptions, and shopping context. Panel III shows two simulation
paradigms, aggregate and individual, with their sales equations. Panel IV shows two causal diagrams
for confounding off and on, which differ only by a dashed arrow from the unobserved shock to price. A
dashed frame around panels III and IV marks the four configurations, and an arrow leads to panel V,
which describes the 32 price scenarios and the counterfactual sales equation.}
\label{fig:generator}
\label{fig:pipeline}

\vspace{3pt}
{\footnotesize\raggedright\emph{Note.} $x$, covariates other than price; $p$, price; $z$, instrumental
variables; $\xi$, the unobserved shock; $\delta$, how strongly price responds to $\xi$,
which creates the confounding; $q$, units sold; $U$, a shopper's utility; $f$, $g$ and $h$, the
generator's price, sales and choice functions.\par}
\end{figure}

\paragraph{Overview.} CausalDemand~\citep{causaldemand} simulates the weekly sales of 40 facial-tissue products in 731 stores over three years, calibrated to seven years of IRI store records. Each dataset comes in two versions: with confounding, discounts deepen when an unobserved demand shock is high; without confounding, they do not depend on it. The generator models shoppers in two ways, a log-linear model of each product's sales and a discrete-choice model of individual shoppers, each in both versions; this paper uses only the log-linear model. Because the generator knows how sales respond to price, it computes the sales at any other price, which serve as the ground truth. Figure~\ref{fig:generator} outlines the generator. \citet{causaldemand} describes it in full, including the data sources, the cost and price processes, the demand equation and its parameter values, and the realism checks.

\paragraph{Data sources.} The generator is built from three kinds of real data, none of which is released: IRI facial-tissue sales from 2001 to 2007, which supply the products, stores, prices, promotion patterns and calibration targets; TDLinx and Spectra store profiles, such as floor space and the number of nearby households, which set each store's sales level; and CDC influenza reports, which set when the cold and flu season peaks in each region. The 40 products are the best sellers of 2007, which account for 80\% of the category's revenue in that year, and the 731 stores are selected in the same way. The first 140 of the 156 weeks are the training weeks, and the last 16, from early September to late December of the third year, are the test weeks.

\paragraph{Prices.}\label{sec:finalprice} A product's typical price is its median price in IRI. The regular price changes in about 4\% of weeks; when it changes, all stores of a chain move to a new price set from the product's current cost. A promotion lowers the prices of all products of one brand in all stores of one chain; promotions last 2.6 weeks on average and are in effect in 23.7\% of weeks, as in IRI. The discount, typically 15 to 35\% of the regular price, is shared by the brand's products, some of which are discounted more deeply than others, and with confounding on it deepens when the product's demand is unusually high. The price paid is rounded to end in 9 cents.

\paragraph{Demand.}\label{sec:demand}\label{sec:loglog} The log-linear model, which builds on that of \citet{magnolfi2025}, sets each product's expected weekly sales in a store from the product's and the store's sales levels, the season, the unobserved shock and the prices of the product and its rivals. Regular prices and discounts enter separately, since shoppers respond differently to a lasting price change and to a temporary discount of the same size. A 1\% increase in the regular price lowers sales by 1.8\% on average, close to the mean of published estimates~\citep{tellis1988}, and a discount raises sales 1.4 times as much as an equal cut in the regular price. The overall sales level is set in each dataset so that a stocked product records no sales in 12\% of weeks, as in IRI.

\paragraph{Confounding and instruments.}\label{sec:endo} The unobserved shock, one for each product, store and week, stands for demand conditions that sales records omit, such as local events or advertising, and shifts sales in every configuration. With confounding on, it also deepens the discount, so that the deepest discounts fall in weeks of high demand. A model that learns this association forecasts well but attributes part of the high demand to the low price, and so overstates the effect of price. The data include two instruments, both generated independently of the shock: the product's cost to the chain when its regular price was last set, which determines the regular price, and a number drawn at the start of each promotion, which sets the depth of the brand's discount. The price changes they cause therefore reflect the effect of price alone, and both are strong instruments~\citep{causaldemand}.

\paragraph{Counterfactual scenarios and paired datasets.}\label{sec:scenario-prices}\label{sec:dgp-repro} A counterfactual scenario changes the discount of the target product or brand in its promoted weeks so that the price paid is 10\% higher or lower, and leaves regular prices, weeks without a promotion and all other products unchanged. The two versions of a dataset are generated from the same random seed and share every random draw, so they differ only in the discounts in promoted weeks, the prices and sales that follow from them, and the overall sales level. The difference between a model's scores on the two versions therefore measures the effect of confounding on that model.

\paragraph{Realism.}\label{app:represent} The simulated prices resemble the real records. Across the four configurations, 35 of 36 measures of how prices move over time, such as the frequency and depth of promotions, lie within the range of the seven IRI years, and a classifier separates simulated from real 13-week price series with 68 to 70\% accuracy, close to its accuracy in separating one IRI year from the other six.

\section{Details on the Experiments}\label{app:eval}\label{sec:release}

\subsection{Data Generation, Splitting and Model Fitting}\label{sec:steps}

This section describes how each simulated dataset is generated, split and supplied to the models. Every seed
and both values of $\delta$ follow the same steps.

\begin{enumerate}[leftmargin=*, itemsep=2pt]
\item \textbf{Simulation.} The simulator generates a panel of 40 products in 731 stores over 156 weeks
with the log-linear demand model (Appendix~\ref{app:data}). For each product, store and week, it draws the cost variables $z$, the
promotion schedule, the discount depth and an unobserved shock $\xi$. It then computes the price $p$
and the expected units sold $q$. With $\delta = 0.30$, the discount depth depends on $\xi$, so price and
$\xi$ are correlated. With $\delta = 0$, they are independent. The two settings share all other random
draws (Appendix~\ref{sec:dgp-repro}). Changing $\delta$ alters the discounts and therefore the share of weeks in which a stocked
product sells nothing. To keep this share at its real-data value, the overall demand level is recalibrated in
each setting (Appendix~\ref{sec:loglog}).

\item \textbf{Temporal split.} Weeks 1--140 form the training set, and weeks 141--156 form the test set.
The training file contains one row for each product, store and week with positive sales. Each row
records the price, the promotion indicator, the product and store attributes and the units sold. No
released file contains $\xi$.

\item \textbf{Model fitting.} We train each model on the training set or supply it as context
(Appendix~\ref{sec:fmts-inputs}). We then freeze the model for both evaluation tasks.
\end{enumerate}

\subsection{Forecasting and Counterfactual Prediction}\label{sec:stack}\label{app:mechanism}

Both evaluations start from the same fitted model and the same held-out rows, every product that a store
stocks in each of the 16 evaluation weeks. In \emph{sales forecasting}, the model predicts the units sold
at the prices the retailer actually charged. In \emph{counterfactual prediction}, it predicts how those
units would change if the price of a product, or of all products of a brand, were raised or lowered by
10\%, with everything else about the store and week held as it was. The two tasks ask for different
quantities. A forecast is the expected sales given the price that was observed, $\mathbb{E}[q \mid p, x]$.
A counterfactual prediction is the expected sales given a price that is set,
$\mathbb{E}[q \mid \mathrm{do}(p), x]$, where $\mathrm{do}(p)$ denotes that the price is assigned rather
than observed. Appendix~\ref{app:truth} describes the price changes and their ground truth, and
Appendix~\ref{sec:layer3} the scores.

\subsection{Counterfactual Scenarios and Ground Truth}\label{app:truth}\label{sec:scenarios}\label{app:scenarios}\label{sec:intervention}

With store, product and week subscripts omitted, the generator sets the price as
$p = f(x, z, \delta\xi)$ and the units sold as $q(p, x, \xi)$. Here $x$ denotes the covariates,
which comprise the product attributes, the store characteristics and the seasonal effects; $z$ the
instruments; $\xi$ the unobserved shock; and $\delta$ the strength with which the price
responds to the shock (0.30 with confounding on and 0 with confounding off). A counterfactual
scenario replaces the actual price $p$ by a hypothetical price $p^{\text{cf}}$ and holds all else
fixed, namely the products, stores and weeks, the covariates $x$, the shock $\xi$ and the random
draws of the generator. On every row of the evaluation file, the ground truth is
\begin{equation*}
q = q(p, x, \xi), \qquad q^{\text{cf}} = q(p^{\text{cf}}, x, \xi), \qquad
\Delta q = q^{\text{cf}} - q,
\end{equation*}
where $q$, the units at the actual price, is the forecasting ground truth and $q^{\text{cf}}$, the
units at the scenario price, is the counterfactual ground truth. Only the price equation is replaced
by the scenario price, $\mathrm{do}(p = p^{\text{cf}})$, and because $q$ and $q^{\text{cf}}$ differ
only in the price, $\Delta q$ is the effect of the price change alone. Computing $q^{\text{cf}}$ requires the
unobserved shock $\xi$, which the input files omit and real sales records never contain; the
generator holds $\xi$ at its actual value, which makes the counterfactual ground truth exact.

The benchmark defines 32 scenarios. Each changes the price of one target, either a single product or
all products of one brand, by $+10\%$ or by $-10\%$, and leaves all other prices unchanged. In every scenario, the ground truth covers every row of the evaluation file, so that it
records both the response of the targets and the shift of sales to and from their rivals. The targets are selected in the
confounding-off dataset and used unchanged in the confounding-on dataset. Only products with sales
in the training windows of both datasets are eligible; a brand scenario moves all of the brand's
eligible products. For each scenario, the
scenario file gives the actual and the scenario price of every row of the evaluation file, and
Appendix~\ref{sec:scenario-prices} describes how the generator applies the price change.

\subsection{Scores for the Two Tasks}\label{sec:layer3}\label{sec:errorpair}\label{sec:conventions}\label{app:metric}

\paragraph{Sales forecasting.}\label{sec:layer1}
The forecast error is the weighted mean absolute percentage error (WMAPE) of a method's predicted
units $\hat q_{sjt}$ at the actual prices, over all rows $(s, j, t)$ of the evaluation file,
\begin{equation*}
\text{forecast error} = \frac{\sum_{s,j,t} |\hat q_{sjt} - q_{sjt}|}{\sum_{s,j,t} q_{sjt}}.
\end{equation*}

\paragraph{Counterfactual prediction.}
The counterfactual scores compare a method's predicted response to a price change with the true
response. They are computed only on the rows in which the scenario changes a target's price, and the
method's predicted percentage change is applied to the true sales. For a scenario with target
products $J$, let $R$ be the rows $(s, j, t)$ with $j \in J$ in which the scenario changes $j$'s price,
and let $\widehat{\Delta q}_{sjt}$ be the method's predicted change. Then
\begin{equation*}
\tilde q^{\text{cf}}_{sjt} = q_{sjt}\Big(1 + \frac{\widehat{\Delta q}_{sjt}}{\hat q_{sjt}}\Big),
\end{equation*}
so that an error in the predicted sales level $\hat q_{sjt}$ cannot offset an error in the predicted
response. The counterfactual error of the scenario is
\begin{equation*}
\text{counterfactual error} = \frac{\sum_R |\tilde q^{\text{cf}} - q^{\text{cf}}|}{\sum_R q^{\text{cf}}}.
\end{equation*}
Rows with $\hat q_{sjt} \le 10^{-9}$ are excluded. The counterfactual error is a WMAPE on the same
scale as the forecast error.

\paragraph{Aggregation.}
For each method and dataset, the 32 scenarios are summarized by the mean counterfactual error. Scores in the paper are means over the five seeds with 95\% $t$-intervals
across seeds.

\subsection{Experiments}\label{sec:fmts-experiments}

Three experiments produce the results of Section~\ref{sec:results}. The first, on real data, asks whether
the models that forecast most accurately also predict that sales fall when price rises
(Table~\ref{tab:actual}). The second and third use the simulated data, on which the counterfactual outcome
is known: the second varies how much data a model receives (Figure~\ref{fig:divergence}), and the third
varies the form in which the instruments reach a model (Table~\ref{tab:results}).
Appendix~\ref{app:models} gives the settings needed to reproduce each.

\subsubsection{Experiment 1: Forecast Accuracy and the Law of Demand on Real Data}\label{sec:exp1}

\paragraph{Question.} Do the models that forecast real sales most accurately also recover the most basic
property of a price response, that a product's sales fall when its own price rises?

\paragraph{Design.} On real data the sales at a price the retailer never set are unobserved, so the
counterfactual error cannot be computed. The direction of the response can still be checked, because a
correct model predicts lower sales after a price increase. We use weekly scanner data of the Dominick's Finer Foods chain for two categories, snack
crackers and breakfast cereals, each with 60 products in 82 or 81 stores over 156 weeks, of which the
last 16 are held out (Appendix~\ref{sec:fmts-table1}). LightGBM and Double ML are
tuned for forecast accuracy, as an analyst would tune them on real data. TabPFN, TabICLv2 and TabFM are run on
five context samples, and their scores are averaged. No model except Double ML receives the instruments.

\paragraph{Measures.} Forecast error is the WMAPE of predicted units against units sold. Own-sign accuracy is the share of rows, in the scenarios that raise a price by 10\%, on which the model predicts that units fall.

\subsubsection{Experiment 2: More Context}\label{sec:exp2}

\paragraph{Question.} Does giving a model more data improve its price response along with its forecast,
and does the answer depend on whether price is confounded?

\paragraph{Design.} Each model is run on training samples of increasing size, from 100 store-weeks to the
full panel of 102,340 store-weeks, up to the largest size it accepts: 6,400 store-weeks for TabPFN and
TabICLv2, 1,600 for CCPFN, 400 for TabFM, and the full panel for LightGBM and Double ML. The samples are nested, so that
each contains every smaller one and the only change from one size to the next is additional rows. No
model except Double ML receives the instruments. The experiment runs on the five datasets with
confounding on (Panel~A) and the five with confounding off (Panel~B). The two datasets of a seed share
every random draw except for the presence of confounding.

\paragraph{Measures.} Forecast and counterfactual error, both from the same runs, as the mean over the five
datasets of a panel with a 95\% interval across datasets. Two reference lines frame the counterfactual
error: predicting no change, and the rounding floor that a model knowing the true demand would still
incur.

\subsubsection{Experiment 3: Three Ways of Supplying the Instruments}\label{sec:exp3}

\paragraph{Question.} Given valid instruments, does supplying them through a model's inputs restore the
causal price response, and does the form in which they are supplied matter?

\paragraph{Design.} Each model runs at one context (Appendix~\ref{sec:fmts-table2}) in three forms that
differ only in the columns added to its inputs: no instruments; control-function residuals; and the raw instruments. The models are those of Experiment~2 together with Chronos-2. Double ML uses the instruments directly and appears once in Table~\ref{tab:results}, as a reference. The datasets are the
ten of Experiment~2.

\paragraph{Measures.} Forecast and counterfactual error, as in Experiment~2.

\FloatBarrier

\section{Implementation Details}\label{app:models}

This appendix gives what is needed to replicate each experiment of Appendix~\ref{sec:fmts-experiments}:
the models (Appendix~\ref{sec:fmts-models}), the settings of each experiment
(Appendix~\ref{sec:fmts-inputs}) and the compute (Appendix~\ref{sec:fmts-compute}).

\subsection{Models and Their Settings}\label{sec:fmts-models}

Table~\ref{tab:models-fmts} lists the implementation and settings of each model.

\begin{table*}[!tp]
\centering
\caption{Models and their settings.}
\label{tab:models-fmts}
\footnotesize
\begin{tabular}{@{}L{0.14\textwidth}L{0.30\textwidth}L{0.50\textwidth}@{}}
\toprule
model & implementation & settings \\
\midrule
\multicolumn{3}{@{}l}{\textbf{Tree model}} \\
LightGBM & \texttt{lightgbm} 4.7.0 & 300 trees; learning rate 0.05; 63 leaves; tuned on the real data (Appendix~\ref{sec:fmts-table1}) \\
\addlinespace[2pt]
\multicolumn{3}{@{}l}{\textbf{Tabular foundation models}} \\
TabPFN & \texttt{tabpfn} 8.5.0, package-default weights & package defaults (8 estimators); fitted with cache \\
TabICLv2 & \texttt{tabicl} 2.2.0, checkpoint \texttt{tabicl-regressor-}\allowbreak\texttt{v2-20260212} & package defaults (8 estimators); key-value cache \\
TabFM & \texttt{tabfm} 1.0.1, weights \texttt{google/tabfm-1.0.0} & 1 estimator; bfloat16 precision \\
\addlinespace[2pt]
\multicolumn{3}{@{}l}{\textbf{Causal foundation model}} \\
CCPFN & \texttt{ccpfn} 0.1.2, weights \texttt{Layer6/CCPFN} & log price as the treatment; context length set to the number of context rows, so that no row is resampled; float32 \\
\addlinespace[2pt]
\multicolumn{3}{@{}l}{\textbf{Time-series foundation model}} \\
Chronos-2 & \texttt{chronos-forecasting} 2.3.1, weights \texttt{amazon/chronos-2} (120M parameters) & zero-shot, no fine-tuning; default context length; median forecast; no learning across series \\
\addlinespace[2pt]
\multicolumn{3}{@{}l}{\textbf{Model with instruments}} \\
Double ML (with instruments) & own implementation; partially linear IV model & LightGBM nuisance models with 200 trees, learning rate 0.05 and 63 leaves; 2-fold cross-fitting; tuned on the real data (Appendix~\ref{sec:fmts-table1}) \\
\addlinespace[2pt]
\multicolumn{3}{@{}l}{\textbf{Reference}} \\
No change & -- & predicts no response to a price change; makes no forecast \\
\bottomrule
\end{tabular}

\vspace{3pt}
{\footnotesize\raggedright\emph{Note.} On the simulated data every model uses the settings above, and
none is tuned. Every model uses random state 42. CCPFN is the released causal foundation model that
accepts a continuous treatment; those for a binary treatment cannot represent a 10\% price change.\par}
\end{table*}

\subsection{Experiment Settings}\label{sec:fmts-inputs}

Each experiment is described under the same headings: data, context, model inputs, tuning, price
changes, and scores and intervals. Experiment~1 gives every setting in full; Experiments~2
and~3 state only what differs.

\subsubsection{Experiment 1: Real Data (Table~\ref{tab:actual})}\label{sec:fmts-table1}

\paragraph{Data.} The snack-cracker and breakfast-cereal datasets of CausalDemand~\citep{causaldemand},
built with the \texttt{causaldemand} package from the Dominick's data of the James M. Kilts Center for
Marketing, University of Chicago Booth School of Business. Weeks 63 to 218 of the Dominick's data are used,
of which weeks 203 to 218 are the test weeks. The datasheet of the benchmark
(\url{https://huggingface.co/datasets/jean-jsj/CausalDemand}) documents their construction, including the
promotion flag and the two instruments.

\paragraph{Context.} The training file contains one row per product, store and week with a
positive sale, and no row for a week without a sale. LightGBM and Double ML are fitted on all training rows.
TabPFN and TabICLv2 receive 6,400 sampled store-weeks and TabFM 400, with sampler seeds 0 to 4. A sample
consists of store-weeks, all rows of one store in one week. The training window's store-weeks are sorted
by store and week, permuted with \texttt{numpy.random.default\_rng}($k$) for sampler seed $k$, and the first $n$ are
kept. Samples of one seed are therefore nested, and every model sees the same rows at a given dataset, size
and seed. Only the training rows are sampled; evaluation always covers every store. Chronos-2 runs once,
on the full 140-week history of each series.

\paragraph{Model inputs.} The six numerical features are the log price paid; the average log price of the
other products in the same store and week, recomputed from whatever prices are supplied, so that a price
change also reaches the other products, and equal to the row's own log price in a store-week with one
product; the promotion flag; a seasonal index, which is constant in the log-linear datasets, since they
release none; the logarithm of one plus the number of households near the store; and the week index,
counted from the first training week. Product, store and brand enter as integer codes. The instruments
are not among the features. All models except Chronos-2 learn the logarithm of units, with units
floored at $10^{-6}$, and forecast the exponential of their prediction, with the prediction clipped to
the interval from $-20$ to 20.

Chronos-2 lays out each product-store series on every training week. A week
without a row has a missing target; its price is taken from the nearest week of the same series, and its
promotion flag from the same chain, brand and week. The numerical features enter as past covariates over
the training weeks and as known future covariates over the evaluation weeks; product, store and brand
are constant within a series and need no covariate. A history of $h$ weeks keeps the last $h$ training
weeks of both the target and the covariates. Each series is forecast on its own, on the logarithm of
units, and the forecast is the exponential of the median forecast.

Double ML needs the regular price and the discount, which the files do not release; they are
reconstructed from the price paid and the promotion flag. On a row without promotion the price paid is the
regular price; on a promoted row the regular price is the nearest observation without promotion of the
same product and store, and the discount is one minus the price paid over the regular price. Double ML
removes the influence of the controls (the promotion flag, the seasonal
index, the household count, the week index and the product, store and brand codes) from log units, from
the four treatments and from the instruments with LightGBM and 2-fold cross-fitting, and estimates the
effects of the treatments by overidentified two-stage least squares on what remains, pooled over products.
The treatments are the log regular price, the discount and the average of each over the other products in
the store and week; the instruments are the log supply cost, the promotion cost with its square and cube,
and the averages of these four over the other products.

\paragraph{Tuning.} LightGBM and Double ML are tuned on each dataset. The first 124 training
weeks fit each of 20 candidates, the default settings and 19 random draws (seed 20261001), and the last 16
training weeks score them by forecast error against realized units; the best candidate is refitted on all
140 weeks. The search ranges are, for LightGBM, 100 to 1,000 trees, learning rate 0.01 to 0.2, 15 to 255
leaves, 5 to 200 rows per leaf, row and feature sampling 0.5 to 1 and L2 penalty 0.001 to 10, and, for the
LightGBM models inside Double ML, 100 to 600 trees, learning rate 0.02 to 0.2, 15 to 127 leaves and 5 to 100
rows per leaf. Ranges are sampled on a log scale, except row and feature sampling, which are sampled
uniformly. The foundation models use the settings of Table~\ref{tab:models-fmts}.

\paragraph{Price changes.} Each fitted model predicts every evaluation row twice, at the
actual prices and at the scenario prices, and the predicted change is the difference. Between the two
predictions, only the price paid and the average price of the other products change; the promotion flag,
the calendar and the store keep their observed values. LightGBM and the tabular foundation models receive
the scenario price as the price paid, and Chronos-2 receives it as a future value of its price covariate.
Double ML applies the scenario's change in the discount through its estimated effects.

\paragraph{Scores and intervals.} Forecast WMAPE is computed over the test rows. Own-sign accuracy uses the
target rows whose price the $+10\%$ scenarios change; a predicted change of exactly zero counts as a failure, and the share is pooled over the scenarios. Such predictions occur on 2.8\% of these rows for LightGBM on cereals and on 0.4\% (crackers) and 1.0\% (cereals) for TabFM, and on none for the other models. The 95\% intervals come from a store-clustered
percentile bootstrap: 1,000 draws of the stores with replacement
(\texttt{numpy.random.default\_rng(20260930)}), each metric recomputed
on the drawn stores with the predictions held fixed, and the 2.5th and 97.5th percentiles over the draws.
One set of draws serves every run; for a sampled model, the value in a draw is the mean over its five runs
in that draw.

\subsubsection{Experiment 2: More Context (Figure~\ref{fig:divergence})}\label{sec:fmts-fig1}

\paragraph{Data.} The five log-linear facial-tissue datasets with confounding on and the five with
confounding off, downloaded from the CausalDemand repository
(\url{https://huggingface.co/datasets/jean-jsj/CausalDemand}), whose datasheet documents every file.

\paragraph{Context.} As in Experiment~1, the training rows are those with a positive sale and samples are
drawn and nested in the same way. LightGBM and Double ML run at 100, 400, 1,600, 6,400 and 25,600
store-weeks and on the full panel; TabPFN and TabICLv2 at 100, 400, 1,600 and 6,400 store-weeks; CCPFN at
100, 400 and 1,600; TabFM at 100 and 400. A sample of 400 store-weeks holds about 9,000 rows, one of 6,400
about 145,000, and the full panel about 2.3 million. Each dataset has one run per model and context, with
sampler seed 0.

\paragraph{Model inputs.} As in Experiment~1, with two differences. In reconstructing the regular price
for Double ML, one cent is added to a price paid that ends in 9 cents, to account for the rounding of prices;
Experiment~1 omits this correction. CCPFN receives the log price as its treatment and the other
features as covariates.

\paragraph{Tuning.} None. Every model uses the settings of Table~\ref{tab:models-fmts}.

\paragraph{Price changes.} The 32 scenarios of the benchmark (Appendix~\ref{app:truth}), applied as in
Experiment~1; CCPFN, like LightGBM and the tabular foundation models, receives the scenario price as the
price paid. Double ML applies the scenario's change in the discount (Appendix~\ref{sec:scenario-prices}).

\paragraph{Scores and intervals.} Scores are computed with the benchmark's
scorer~\citep{causaldemand}, which implements Appendix~\ref{sec:layer3}. A reported value is the mean over
the five datasets of a panel with the half-width of the 95\% $t$-interval, $t_{0.975,4}\,s/\sqrt{5}$,
where $s$ is the standard deviation across the datasets, so the interval includes the variation due to
sampling the context. No change ($\Delta\hat q = 0$) is scored like a model on every dataset. The
rounding floor is the benchmark's released reference, which applies the true demand to the scenario
prices as recorded. Recorded prices are rounded to end in 9, whereas demand responds to the unrounded
price (Appendix~\ref{sec:finalprice}), so even the true demand incurs this error.

\subsubsection{Experiment 3: Three Ways of Supplying the Instruments
(Table~\ref{tab:results})}\label{sec:fmts-table2}\label{app:forms-input}

\paragraph{Data.} As in Experiment~2.

\paragraph{Context.} LightGBM and Double ML on the full panel; TabPFN and TabICLv2 on 6,400 store-weeks;
TabFM and CCPFN on 400; Chronos-2 on the 140 weeks of each series. The no-instrument cells of LightGBM and the
tabular foundation models are the runs of Experiment~2 at these contexts.

\paragraph{Model inputs.} As in Experiment~2, with the instruments added in one of three forms: none, the
residuals or the raw instruments. The residuals are built as follows. The four price components of
Experiment~1, the log regular price, the discount and the average of each over the other products in the
store and week, are each regressed by least squares on the eight instrument columns below, the promotion
flag, the household count, the week index, product indicators and an intercept (the seasonal index is
dropped as constant). The regression is fitted on the training rows of the run, so it is shared by every
model at the same dataset, size and sampler seed. The residual of a row is its reconstructed component
minus the fitted value at its instruments and controls; evaluation rows use the components reconstructed
from the actual evaluation prices. The raw instruments are eight columns: the log supply cost, the
promotion cost with its square and cube, and the average of each of these four over the other products in
the store and week. No first stage is fitted. LightGBM and the tabular foundation models receive the
residuals or the raw instruments as additional features. CCPFN receives one layout of covariates in every
form, with the columns that a form does not carry set to zero. Chronos-2 receives them as additional
covariate series, past and future, as it receives the price; no layer is added and no weights are trained.

\paragraph{Tuning.} None, as in Experiment~2.

\paragraph{Price changes.} As in Experiment~2. The residuals and the raw instruments keep their observed
values under a price change; a residual stands in for the unobserved shock, which the scenario holds fixed.

\paragraph{Scores and intervals.} As in Experiment~2.

\subsection{Compute}\label{sec:fmts-compute}

The foundation models ran on cloud GPUs, an NVIDIA A10 with 24 GB, except TabICLv2 at 6,400 store-weeks,
which needs an 80 GB A100 on the simulated data and an H200 on the real data; the GPU does not change the
settings. LightGBM and Double ML ran on CPUs, with Python 3.14.2, \texttt{numpy} 2.5.2, \texttt{pandas} 3.0.5 and
\texttt{scikit-learn} 1.9.0. The foundation-model runs used Python 3.12.10 with
\texttt{torch} 2.13.0, \texttt{numpy} 2.5.2, \texttt{pandas} 3.0.5 and \texttt{scikit-learn} 1.9.0, except
CCPFN, which used \texttt{torch} 2.5.1, \texttt{numpy} 2.2.6 and \texttt{scikit-learn} 1.9.1. In timed runs, a run took about 5.6 minutes for TabPFN and 2.9 for
TabICLv2 at 6,400 store-weeks, 1.3 for CCPFN at 400 and about 21 at 1,600, 1.9 to 3.1 for Chronos-2, and 1.0 for LightGBM
and 6.9 for Double ML on the full panel.

\FloatBarrier

\end{document}